\documentclass[11pt,a4paper]{article}
\usepackage[hyperref]{emnlp2018}
\usepackage{times}
\usepackage{latexsym}
\usepackage{url}

\usepackage{amsmath}
\usepackage{amssymb}
\usepackage{mathtools}
\usepackage[shortcuts]{extdash}
\usepackage{booktabs}
\usepackage{tabularx}
\usepackage{subcaption}
\usepackage{graphicx}
\usepackage{placeins}
\usepackage{float}

\newcommand{\R}{\mathbb{R}}

\newcommand{\z}{\mathbf{z}}

\newcommand{\e}{\mathbf{e}}

\newcommand{\bfat}{\mathbf{b}}
\newcommand{\uf}{\mathbf{u}}

\newcommand{\p}{\mathbf{p}}

\newcommand{\E}{\mathbf{E}}

\newcommand{\Zf}{\mathbf{Z}}
\newcommand{\U}{\mathbf{U}}

\newcommand{\V}{\mathbf{V}}

\newcommand{\sparsemax}{\text{Sparsemax}}
\newcommand{\transpose}{^\mathsf{T}}

\newenvironment{tightcenter}{%
  \setlength\topsep{0pt}
  \setlength\parskip{0pt}
  \begin{center}
}{%
  \end{center}
}

\renewcommand{\epsilon}{\ensuremath\varepsilon}

\renewcommand{\phi}{\ensuremath{\varphi}}

\newcommand\citecomment[3]{\citep[#1][#3]{#2}}

\newcommand\cbraces[1]{\{#1\}}

\aclfinalcopy 

\title{Learning and Evaluating Sparse Interpretable Sentence Embeddings}

\author{Valentin Trifonov \\
  ETH Z{\"u}rich, Switzerland \\
  {\tt valentin.trifonov@outlook.com} \enspace \\\And
  Octavian-Eugen Ganea \\
  ETH Z{\"u}rich, Switzerland \\
  {\tt \enspace octavian.ganea@inf.ethz.ch} \\\AND
  Anna Potapenko \\
  National Research University \\
  Higher School of Economics, Russia\footnotemark \\
  {\tt anna.a.potapenko@gmail.com} \\\And
  Thomas Hofmann \\
  ETH Z{\"u}rich, Switzerland \\
  {\tt thomas.hofmann@inf.ethz.ch} \\}

\date{}

\begin{document}
\maketitle
\begin{abstract}
    Previous research on word embeddings has shown that sparse representations, which can be either learned on top of
    existing dense embeddings or obtained through model constraints during training time, have the benefit of increased
    interpretability properties: to some degree, each dimension can be understood by a human and associated with a
    recognizable feature in the data.
    In this paper, we transfer this idea to sentence embeddings and explore several approaches to obtain a sparse
    representation.
    We further introduce a novel, quantitative and automated evaluation metric for sentence embedding interpretability,
    based on topic coherence methods.
    We observe an increase in interpretability compared to dense models, on a dataset of movie dialogs and on the scene
    descriptions from the MS COCO dataset.
\end{abstract}

\section{Introduction}

\renewcommand*{\thefootnote}{\fnsymbol{footnote}}
\footnotetext[1]{Work done during an internship at ETH Z{\"u}rich.}
\renewcommand*{\thefootnote}{\arabic{footnote}}
In the word embeddings literature, it has previously been of interest to find interpretable representations:
individual dimensions should capture a distinct semantic meaning, such that humans are able to understand why a word is
encoded in a particular vector.
With a cognitive plausibility argument from~\citet{nnse}, the interpretability can be linked to sparse representations:
they argue that the representation should model a wide range of features in the
data and that every sample should be characterized by the presence of a small number of key features.
\citet{arora}~use this idea to recover and disentangle the different meanings of polysemous words.

The above\-/named approaches, as well as those by~\citet{spine,spowv}, recover an interpretable sparse representation
in a separate, post-processing step on top of the uninterpretable, dense embeddings of the original model (often
word2vec or GloVe).
This is commonly done using sparse coding or a downstream model.
Additionally to understanding a model's intermediate representation, there has been work
on constructing models that inherently use a sparse embedded representation by learning it during the training
process~\citep{sun2016sparse,chen2017learning}.
This is motivated by the idea that the model should include the prior that each word is a sparse combination of
disentangled features from the very beginning.
In contrast, when computing dense embeddings first, it is less likely that this representation will be easily
disentanglable in the post-processing step.

\citet{thoughtvectors} argues that sparse representations can be used to explain image and sentence embeddings as well.
To be precise, the author focuses on encoder-decoder neural networks and uses sparse coding to recover interpretable
features in the latent spaces of a variational autoencoder~\citep{vae} and an image captioning system based
on~\citep{showandtell}.

In this paper, we aim to use sparse methods to disentangle sentence embeddings' dimensions.
We focus on a simple sentence autoencoder model, and apply both a sparse-coding-based post-processing
technique, as well as model constraints during training time, to obtain sparse vector representations of sentences.
We aim to increase the understanding of the latent space, which helps us gain insight into how the inference and
learning process works by identifying the patterns in the data that the model learns to recognize and
encode in this representation.

To compare our different approaches, as well as measure the improvement compared to the baseline of a dense autoencoder model,
we introduce a novel, quantitative and automated metric of the mentioned interpretability properties.
It is based on the notion of topic coherence and further develops it for the case of sentences.
We observe that the new measure reflects our manual judgment on the interpretability of the embeddings.
Additionally, we track reconstruction quality and performance in downstream tasks, showing that sparse approaches
can obtain a remarkable increase in interpretability at a moderate cost in quality.

\section{Models}
\label{sec:models}

Our models are based on a standard recurrent neural network autoencoder following the \emph{Sequence-to-Sequence}
architecture~\citecomment{seq2seq;}{seq2seq}{}.
This architecture is based on the encoder-decoder scheme, where an encoder network maps the input to a dense, embedded
representation $\z$, and a decoder net reconstructs the input from $\z$.
In Section~\ref{subsec:experiments}, we give a more detailed description of our experimental setup.

\subsection{Enforcing Sparsity by Post-Processing Dense Embeddings}

Consider a dataset $x_1, \ldots, x_N$ of $N$ sentences.
We train a dense autoencoder net with a hidden state size $D'=500$ to convergence, and compute
$\Zf = [\z_1, \ldots, \z_N]\transpose \in \R^{N \times D'}$, the vector representations of the data arranged as
the rows of a matrix.
We follow the approaches by~\citet{arora,thoughtvectors} and compute a sparse representation of size $D=2000$
on top of $\Zf$, where all but $k$ values have to be zero.
We do this by solving the following sparse dictionary learning problem:
\begin{equation}
    \begin{split}
        \E, \U = \arg \min_{\E, \U} ||\E \U - \Zf ||_F^2, \\
        s.t. \; ||\e_{i}||_0 \leq k, ||\uf_{j}||_2 = 1, \; \forall i, j,
    \end{split}
\end{equation}
whereby we obtain $\E = [\e_1, \ldots, \e_N]\transpose \in \R^{N \times D}$, a set of new, sparse vector representations,
and $\U = [\uf_1, \ldots, \uf_D]\transpose \in \R^{D \times D'}$, a dictionary of \emph{atoms} found in $\Zf$.
We solve this problem with the $k$-SVD algorithm~\citecomment{}{ksvd}{%
we use an open source implementation called \emph{pyksvd}\footnote{\url{https://github.com/hoytak/pyksvd}}}.

The intuition behind this sparse coding approach is as follows.
The atoms $\U$ are intended to represent a wide range of the 2000 most important features that explain the data in the
dense latent space of the model.
By solving this problem we decompose the intermediate representation $\z_i$ of a sample $x_i$ into a linear combination
$\e_i$ of atoms.
By constraining $\e_i$ to a fixed and low sparsity level $k$ we aim to disentangle this representation and therefore
increase interpretability.
We refer to this representation as the $k$-SVD model.

\subsection{Enforcing Sparsity during Embedding Learning}

The $k$-SVD model proposed in the previous section obtains sparse representations through solving two independent
problems: finding a fixed-size vector representation for sentences with a neural model and, in a separate step, mapping
it to an interpretable, sparse representation.
As we mention in the introduction, we conjecture that interpretability can be further increased with an end-to-end
approach.
In this section, we introduce modifications to the model architecture that will force the neural nets to encode and
understand sparse representations of the data during training time.

We propose an additional layer that is inserted between the encoder and decoder net.
We map the vector representation $\z$ to a vector $\e$ of the same size in a \emph{sparsity transformation}.
The only requirements for this mapping are that the output $\e$ is sparse and differentiable (or that we can
define a custom gradient) to allow backpropagation through it.
We then feed $\e$ through the decoder net instead of $\z$ and train the whole net end-to-end.
In the rest of this section, we propose two mappings for such a sparsity transformation.

\subsubsection*{$k$-Sparse.}
For this model, we draw inspiration from the \emph{$k$-Sparse Autoencoder} by~\citet{ksparseae}.
We again introduce a hyperparameter $k$ and define $\e$ by keeping the $k$ largest activations in $\z$,
the \emph{support set}, and setting all other units to zero.
We backpropagate only through the support set.
This is a simple way of enforcing a hard specification for the sparsity level as an integral part of the model.

\subsubsection*{Sparsemax.}
The $k$-Sparse approach has the drawback of requiring a fixed sparsity level for all samples.
To allow for a variable, per-sample sparsity level, we use the \emph{Sparsemax} layer, introduced by \citet{sparsemax}.
Sparsemax is an alternative to Softmax---however, unlike Softmax, it is able to return sparse probability distributions.
It is defined as:
\begin{equation}
    \sparsemax(\z) = \arg\min\limits_{\p \in \Delta^{D - 1}} ||\z - \p||_2^2,
\end{equation}
where $D$ is the dimensionality of $\z$ and $\Delta^{D - 1}$ is the $(D-1)$-dimensional simplex
$= \cbraces{\p \in \R^D \, | \, \mathbf{1}\transpose \p = 1, \p \geq 0}$.

Similar to Softmax, Sparsemax supports a temperature mechanism, where a hyperparameter
$\tau$ trades off the ``confidence'' in the output probability of the largest input unit.
To be precise, as $\tau$ approaches $0$, the probability distribution $\sparsemax\left(\frac{\z}{\tau}\right)$
approaches the distribution peaked on the maximum components of $\z$.
Additionally to Softmax, Sparsemax has the property that this output distribution becomes increasingly sparse.

Putting this together, we introduce a hyperparameter $\tau$ and define a sparsity transformation by defining
$\e = \sparsemax\left(\frac{\z}{\tau}\right)$.

\section{Experiments}
\label{subsec:experiments}

\subsection{Training Details}

In our experiments, we use a vocabulary size of 20,000, with the symbolic words \texttt{<person>}, \texttt{<unk>}, and
\texttt{<eos>} for names and out-of-vocabulary (OOV) words in the dataset, and the end-of-sentence marker, respectively.
We convert words to 100-dimensional word embeddings by looking them up in a trainable matrix $\V$ (note that, in
general, this matrix is not sparse---sparsity is only imposed on the latent space of the sentences\footnote{%
On a side note, sparsity can be imposed on the word embeddings by adding an L1\-/regularizer to
$\V$~\citep{sun2016sparse,chen2017learning}.}).

Our encoder and decoder nets are recurrent neural networks that use a single GRU~\citep{gru} layer.
They have the same hidden dimensionality but share no parameters.
We obtain the model predictions as the Softmax of a learned, affine transformation to 20,000-dimensional space at every
time step of the decoder net.
We minimize the mean cross-entropy loss over all timesteps.
We use a batch size of 64 and the Adam optimization algorithm.

\subsection{Data}

We train our models on the Cornell Movie-Dialogs Corpus and MS Common Objects in Context
datasets~\citecomment{respectively}{cornell,coco}{}.

The Movie-Dialogs Corpus is a collection of movie lines, therefore it contains a wide variety of different utterences
and allows us to explore general-purpose sentence embeddings.
We preprocess this data by splitting the movie lines into separate sentences, thereby obtaining more than 500,000
samples.
This dataset has no predefined split;
we define a validation and test set by setting aside 50,000 samples each.

The MS COCO dataset contains images showing scenes with objects in numerous configurations.
Every image contains 5 human-annotated variations of a caption that describe the scene.
In our experiments, we use only these captions and refer to this as the COCO Captions data.
They total over 600,000 samples and allow us to explore sentence embeddings of a more narrow language:
since they merely describe objects and scenes, they tend to follow the same, simple sentence structure.
The dataset comes with a predefined training\slash validation split.

For tokenizing and splitting movie lines into sentences we use the NLP library SpaCy\footnote{\url{https://spacy.io}}.
All our models are implemented in TensorFlow~\citep{tensorflow}.

\begin{table*}[t!]
    \small
    \centering
    \begin{tabularx}{\textwidth}{l | X}

        original & \texttt{a room with blue walls and a white sink and door .} \\
        reconstruction & \texttt{a room with blue walls and a white sink and windows . } \\
        \hline

        original & \texttt{two cars parked on the sidewalk on the street} \\
        reconstruction & \texttt{two buses parked on the curb on the street } \\
        \hline

        original & \texttt{two women waiting at a bench next to a street .} \\
        reconstruction & \texttt{two women sit at a park next to a street . } \\
        \hline

        original & \texttt{a car that seems to be parked illegally behind a legally parked car} \\
        reconstruction & \texttt{a car that seems to be parked close to a police officer and talking } \\
        \hline

        original & \texttt{a bathroom sink and various personal hygiene items .} \\
        reconstruction & \texttt{a bathroom sink and various other hygiene items . } \\
        \hline

        original & \texttt{this is an open box containing four cucumbers .} \\
        reconstruction & \texttt{this is an open box makes delicious doughnuts . } \\

    \end{tabularx}
    \caption{Typical sentence reconstruction errors by the $k$-Sparse, $k=15$ model, trained on the COCO Captions data.}
    \label{tab:samplesentences}
\end{table*}

\section{A Quantitative and Automated Evaluation Metric}
\label{sec:evaluation}

The most common quantitative interpretability measure for embeddings (in particular word embeddings) is
the intrusion test, first introduced in \citep{wordintrusion}.
This test involves generating 5-tuples of samples, where according to the embeddings model four are related and one
stands out.
The better human judges identify the intruder, the more interpretable the model is considered.

This evaluation method has the drawback of requiring human attention, thereby it is expensive and slow to evaluate.
For our evaluation, we introduce an automated interpretability test, based on topic coherence, that does not require
human attention.
We describe our method in this section.

A \emph{topic model} defines a set of topics in a corpus of documents and allows us to find the top $n$ most likely words
that belong to each topic.
\emph{Topic coherence} is an automated evaluation method of the interpretability of topic models, which has been shown to
correlate well with human assessments~\citep{topiccoherence,mimno2011optimizing}.
Given a symmetric similarity measure of two words (e.g.\ pointwise mutual information),
the coherence of a topic is defined as the mean pairwise similarity of all pairs of words.
The total topic coherence of the model is the mean coherence over all topics.

We devise an evaluation scheme based on topic coherence.
Instead of looking at words in topics, we consider the highest-ranked sentences in the dimensions of our embeddings and replace
the word similarity measure with a sentence similarity measure.
Let $x_d^{(p)}$ be the sample that has rank $p$ in the order given by the $d$-th dimension in the embedding.
For a similarity measure $sim_*$, the coherence of a single dimension $d$ is defined as:
\begin{equation}
    \begin{multlined}
        coh_*(d) = \frac{2}{n \cdot (n-1)} \sum_{p=1}^{n-1} \sum_{q=p+1}^{n} sim_*(x_d^{(p)},x_d^{(q)}).
        \label{eq:tcformula}
    \end{multlined}
\end{equation}
The coherence of the model is defined as the mean coherence over all dimensions:
\begin{equation}
    coh_*(1, \ldots, D) = \frac{1}{D} \sum_d coh_*(d).
    \label{eq:tcformulaall}
\end{equation}
In addition, to determine how much the coherence deteriorates when looking beyond the top ranks, we consider all non-zero samples
of a dimension and we evaluate Equation~\ref{eq:tcformula} on $n$ sentences sampled at random and without replacement from
$\cbraces{x_i \mid e_{i,d} \neq 0}$ instead of $x_d^{(1)}, \ldots, x_d^{(n)}$.

We compute this on the validation set of our data.
We strip all stop words from all sentences.
We consider $n=10$ sentences per dimension, unless a dimension has a non-zero value for less than $n$ samples, in which
case we compute Equation~\ref{eq:tcformula} on all pairs of sentences.
In the following, we define three choices for a sentence similarity measure $sim_*$.

\paragraph{Jaccard Similarity.}
We regard the sentences as sets of words and compute the Jaccard similarity:
\begin{equation}
    sim_\text{J}(x_i, x_j) = \frac{|x_i \cap x_j|}{|x_i \cup x_j|}.
\end{equation}

\paragraph{BoW Similarity.}
We consider the Bag-of-Words (BoW) vectors $\bfat_i, \bfat_j$ of the two sentences, i.e.\ the vectors with the number of
occurrences of each vocabulary word in $x_i, x_j$, respectively.
The similarity is defined as the cosine of the angle between these vectors:
\begin{equation}
    sim_{\text{BoW}}(x_i, x_j) = \frac{\bfat_i\transpose \bfat_j}{||\bfat_i||_2 \, ||\bfat_j||_2}.
\end{equation}

\paragraph{WMD Similarity.}
The Jaccard and BoW similarity measures have a drawback in that they do not take semantic relatedness of different
words into account.
The \emph{Word Mover's Distance} \citecomment{WMD;}{wmd}{} remedies this problem:
the authors define a document distance measure that relies on the word2vec latent space to make a better assessment
of the semantic distance of sentences, based on the distance of the words they consist of.
We use the negative WMD to obtain a similarity measure:
\begin{equation}
    sim_{\text{WMD}}(x_i, x_j) = -\text{WMD}(x_i, x_j).
    \label{eq:wmdsimilarity}
\end{equation}

\section{Results}
\label{sec:results}

\subsection{Reconstruction Quality}

\begin{table*}[t!]
    \small
    \centering

    \begin{subtable}{\linewidth}
        \begin{tabularx}{\textwidth}{c X}
            $e_{i,d_1}$ & $x_i$ \\
            \hline
            $1.00$ & \texttt{a person laying on a couch with a laying on him} \\
            $1.00$ & \texttt{a cat laying on top of a suitcase laying on the floor .} \\
            $1.00$ & \texttt{a man laying on top of a sandy beach laying next to a surfboard .} \\
            $1.00$ & \texttt{a person laying on a couch with a cat laying in their arms, covering part of the face .} \\
            $1.00$ & \texttt{a woman is laying on a couch with a boy laying his head on her belly, and a cat between her legs .} \\
            $1.00$ & \texttt{some cats laying on a dock with their chins laying over the end} \\
            $1.00$ & \texttt{a man laying in bed with a gray cat laying on top of him .} \\
            $1.00$ & \texttt{a number of cows laying in a lot near cars} \\
            $1.00$ & \texttt{a number of items laying on a surface near one another} \\
            $1.00$ & \texttt{two cows laying out together underneath a tree .} \\
        \end{tabularx}
        \caption{This dimension clearly corresponds to sentences that describe a configuration of an object
            \texttt{laying} on another, whether that be people on the couch or items on a surface.
            Coherence score: $coh_\text{WMD}(d_1) = -2.32$.
        }
    \end{subtable}
    \medskip

    \begin{subtable}{\linewidth}
        \begin{tabularx}{\textwidth}{c X}
            $e_{i,d_2}$ & $x_i$ \\
            \hline
            $0.94$ & \texttt{a motorcycle parked outside the doors of a building} \\
            $0.94$ & \texttt{a blue motorcycle parked outside of a building .} \\
            $0.94$ & \texttt{traffic lights on the road showing the street} \\
            $0.93$ & \texttt{food in a bowl sitting on a table} \\
            $0.92$ & \texttt{a yellow train in an outside train station .} \\
            $0.92$ & \texttt{a motorcycle sits on a sidewalk near a building} \\
            $0.92$ & \texttt{a car that is outside in the dirt .} \\
            $0.92$ & \texttt{a red truck parked outside in the snow .} \\
            $0.91$ & \texttt{a boy sitting on a bench at the park} \\
            $0.91$ & \texttt{a black motorcycle is parked on a sidewalk} \\
        \end{tabularx}
        \caption{This dimension seems to capture, with some false positives, different kinds of motor vehicles
            (\texttt{motorcycle}, \texttt{train}, \texttt{car}, \texttt{truck}) that are \texttt{parked}
            (\texttt{sit}, \texttt{sitting}, \texttt{is outside}) somewhere.
            Coherence score: $coh_\text{WMD}(d_2) =  -2.83$.
        }
    \end{subtable}
    \medskip

    \begin{subtable}{\linewidth}
        \begin{tabularx}{\textwidth}{c X}
            $e_{i,d_3}$ & $x_i$ \\
            \hline
            $0.79$ & \texttt{herd of goats in grassy area with herder .} \\
            $0.64$ & \texttt{herd of five zebras grazing in a field} \\
            $0.63$ & \texttt{people are sitting in lounge chairs on the beach .} \\
            $0.63$ & \texttt{a close up of many large kites near the ground} \\
            $0.63$ & \texttt{cows lounge in a field with a mountain backdrop .} \\
            $0.62$ & \texttt{close up of the flower extending from a banana tree stalk} \\
            $0.61$ & \texttt{a group of object on top of a muddy river .} \\
            $0.61$ & \texttt{many plants and umbrellas on the side of the street .} \\
            $0.58$ & \texttt{a close up view of sheets that are on a bed} \\
            $0.58$ & \texttt{room with cramped quarters holding dining table set and extra chairs .} \\
        \end{tabularx}
        \caption{It is not clear which features this dimension captures.
            Coherence score: $coh_\text{WMD}(d_3) =  -3.12$.
        }
    \end{subtable}

    \caption{Examples of selected dimensions $d_1, d_2, d_3$ of our $k$-Sparse, $k=15$ model, trained on the COCO
        Captions data.
        We show the 10 highest-ranked samples $x_i$ and the coherence $coh_\text{WMD}(d)$
        of each dimension $d$.
        We give more examples of high-coherence dimensions in the appendix, in Table~\ref{tab:appendixtopsamples}.
    }
    \label{tab:rnntopsamples}
\end{table*}

We start off by looking at the amount of information lost by our models due to sparsity constraints.

In general, we observe that as the sparsity level is decreased, the reconstructions start to deteriorate.
At low values of $k$, our sparse models often fail to restore the exact meaning or phrasing, but still generate
sentences with correct grammar and related topics.
For example, they turn ``\texttt{waiting at a bench}'' into ``\texttt{sit at a park}'', ``\texttt{sidewalk}'' into
``\texttt{curb}'', ``\texttt{two cars}'' into ``\texttt{two buses}'', and similar.
The $k$-SVD model generally does this less than the other models but in some cases it fails as well.
See examples of typical reconstruction errors by our $k$-Sparse, $k=15$ model in Table~\ref{tab:samplesentences}.

\subsection{Highest-Ranked Samples}
\label{subsec:highestrankedsamples}

We examine the top samples in the dimensions of our embedding models and observe that sparse models often group
sentences s.t.\ they have a common syntactic element or talk about a common concept.
For example, in our $k$-Sparse, $k=15$ model trained on the COCO Captions dataset, we identify dimensions that
represent sentences about objects in \texttt{water}, people \texttt{holding} things, \texttt{horse} (and occasionally
\texttt{bicycle}) \texttt{riders}, sentences starting with common prefixes such as \texttt{there is a [\ldots]}, etc\@.
We give examples in Table~\ref{tab:rnntopsamples} and in the appendix in Table~\ref{tab:appendixtopsamples}.
For some dimensions, this pattern is not only recognizable in the top ranks but for all samples $x_i$ with
$e_{i,d} \neq 0$.

We are able to find such patterns in all sparse models, but the lower the sparsity level, the more apparent these
patterns become.
$k$-SVD based models exhibit these properties to a lesser extent.

\begin{table*}[h!]
    \centering
    \footnotesize

    \begin{subfigure}{\textwidth}
            \begin{tabular}{l r r r}
        Embeddings model & Jaccard & BoW & WMD \\
        \hline
        COCO Captions & $0.05$ & $0.10$ & $-3.12$ \\
        Movie-Dialogs & $0.08$ & $0.16$ & $-2.06$ \\
    \end{tabular}

        \centering
        \caption{Mean similarity of random sentences}
    \end{subfigure}
    \bigskip

    \begin{subfigure}{\textwidth}
        \centering
            \begin{tabular}{l r r r r r r}
        & \multicolumn{3}{c}{Top 10 samples} & \multicolumn{3}{c}{Random 10 samples} \\
        Embeddings model & Jaccard & BoW & WMD & Jaccard & BoW & WMD \\
        \hline
        dense, 500 dim. AE & $0.08$ & $0.14$ & $-3.00$ & $0.06$ & $0.10$ & $-3.12$ \\
        \hline
        $k$-SVD, $k=100$ & $0.07$ & $0.12$ & $-3.08$ & $0.06$ & $0.10$ & $-3.11$ \\
        $k$-SVD, $k=50$ & $0.08$ & $0.13$ & $-3.03$ & $0.06$ & $0.11$ & $-3.10$ \\
        $k$-SVD, $k=20$ & $0.11$ & $0.18$ & $-2.88$ & $0.06$ & $0.11$ & $-3.08$ \\
        $k$-SVD, $k=15$ & $\mathbf{0.11}$ & $\mathbf{0.19}$ & $-2.86$ & $0.07$ & $0.12$ & $-3.06$ \\
        \hline
        $k$-Sparse, $k=100$ & $0.08$ & $0.14$ & $-3.02$ & $0.06$ & $0.11$ & $-3.09$ \\
        $k$-Sparse, $k=50$ & $0.09$ & $0.15$ & $-2.96$ & $0.07$ & $0.12$ & $-3.06$ \\
        $k$-Sparse, $k=20$ & $0.11$ & $0.17$ & $\mathbf{-2.85}$ & $0.08$ & $0.14$ & $\mathbf{-3.00}$ \\
        $k$-Sparse, $k=15$ & $0.11$ & $0.18$ & $-2.86$ & $\mathbf{0.08}$ & $\mathbf{0.14}$ & $-3.01$ \\
        Sparsemax, $\tau=50$ & $0.04$ & $0.07$ & $-3.25$ & $0.03$ & $0.06$ & $-3.27$ \\
        Sparsemax, $\tau=20$ & $0.04$ & $0.06$ & $-3.29$ & $0.03$ & $0.05$ & $-3.35$ \\
        Sparsemax, $\tau=10$ & $0.04$ & $0.07$ & $-3.25$ & $0.03$ & $0.06$ & $-3.31$ \\
    \end{tabular}

        \caption{COCO Captions dataset}
        \label{tab:tcevaluationcoco}
    \end{subfigure}
    \bigskip

    \begin{subfigure}{\textwidth}
        \centering
            \begin{tabular}{l r r r r r r}
        & \multicolumn{3}{c}{Top 10 samples} & \multicolumn{3}{c}{Random 10 samples} \\
        Embeddings model & Jaccard & BoW & WMD & Jaccard & BoW & WMD \\
        \hline
        dense, 500 dim. AE & $0.20$ & $0.31$ & $-1.85$ & $0.09$ & $0.16$ & $-2.02$ \\
        \hline
        $k$-SVD, $k=100$ & $0.17$ & $0.24$ & $-1.99$ & $0.09$ & $0.16$ & $-2.01$ \\
        $k$-SVD, $k=50$ & $0.17$ & $0.24$ & $-2.01$ & $0.10$ & $0.16$ & $-2.01$ \\
        $k$-SVD, $k=20$ & $0.20$ & $0.28$ & $-1.91$ & $0.11$ & $0.18$ & $-2.01$ \\
        $k$-SVD, $k=15$ & $0.20$ & $0.29$ & $-1.88$ & $0.12$ & $0.19$ & $-1.98$ \\
        \hline
        $k$-Sparse, $k=100$ & $0.16$ & $0.25$ & $-2.01$ & $0.10$ & $0.18$ & $-2.08$ \\
        $k$-Sparse, $k=50$ & $0.16$ & $0.25$ & $-1.95$ & $0.11$ & $0.19$ & $-2.05$ \\
        $k$-Sparse, $k=20$ & $0.19$ & $0.30$ & $-1.82$ & $0.13$ & $0.22$ & $-1.99$ \\
        $k$-Sparse, $k=15$ & $\mathbf{0.22}$ & $\mathbf{0.33}$ & $\mathbf{-1.76}$ & $0.14$ & $\mathbf{0.23}$ & $-1.98$ \\
        Sparsemax, $\tau=50$ & $0.12$ & $0.19$ & $-2.13$ & $0.12$ & $0.19$ & $-2.02$ \\
        Sparsemax, $\tau=20$ & $0.13$ & $0.21$ & $-2.02$ & $\mathbf{0.16}$ & $0.23$ & $\mathbf{-1.89}$ \\
        Sparsemax, $\tau=10$ & $0.15$ & $0.22$ & $-2.01$ & $0.15$ & $0.22$ & $-1.96$ \\
    \end{tabular}

        \caption{Movie-Dialogs dataset}
        \label{tab:tcevaluationcornell}
    \end{subfigure}

    \caption{Interpretability of our models, as measured by our topic-coherence-based metric in
        Equation~\ref{eq:tcformulaall}.
        We evaluate this equation using three different notions of sentence similarity $sim_*$.
        In Equation~\ref{eq:tcformula}, we consider 10 random non-zero samples in addition
        to the 10 highest-ranked samples.}
    \label{tab:tcevaluation}
\end{table*}

\subsection{Quantitative Evaluation}

In Table~\ref{tab:rnntopsamples} we additionally report the coherence $coh_\text{WMD}(d)$ of the presented dimensions $d$
(see Equations~\ref{eq:tcformula},~\ref{eq:wmdsimilarity}).
We observe that this score correlates with our empirical assessment of the interpretability of the dimension.
For example, we observe on the COCO dataset that, while unrelated groups of sentences usually have a coherence score
of $< -3$, sentences with common or semantically related subjects and objects have higher coherence scores (usually
between $-2.8$ and $-2.2$).
Groups of sentences with very close semantic meaning or large common prefixes have coherence scores around $-2$ or
higher.

We report the topic coherence of our models (Equation~\ref{eq:tcformulaall}) in
Table~\ref{tab:tcevaluation}.
As rough reference values for the metrics, we include the mean similarity of pairs of random sentences from the
dataset (estimated on 500 randomly sampled pairs), and the topic coherence of a 500-dimensional dense
autoencoder model.

In accordance with our empirical observations, we see an increase in interpretability in the sparse models.
For example, on the COCO Captions data, a random pair of sentences has a WMD-based similarity of -3.12,
and the WMD-based coherence score of a dense autoencoder model is -3.
With the additional sparse coding step on top of that, we can increase the coherence to -2.86.

\subsection{Downstream Tasks}
\label{sec:downstream}

\begin{table*}
    \footnotesize
    \centering

    \begin{subfigure}{\textwidth}
        \centering
        \setlength{\tabcolsep}{4pt}
            \begin{tabular}{l@{\hspace{6pt}}r r r r r r r r r r r }
        Embeddings model & CR & MR & SUBJ & MPQA & SST2 & SST5 & TREC & SICK-E & SICK-R & STS14 & MRPC \\
        \hline
        dense, 500 dim. AE & $65.99$ & $\mathbf{59.37}$ & $76.24$ & $\mathbf{73.01}$ & $60.63$ & $28.96$ & $\mathbf{77.60}$ & $\mathbf{75.50}$ & $\mathbf{0.61}$ & $\mathbf{0.42}$ & $67.25$ \\
        \hline
        $k$-SVD, $k=100$ & $60.48$ & $54.63$ & $69.61$ & $70.73$ & $59.58$ & $25.07$ & $68.20$ & $56.36$ & $0.34$ & $0.18$ & $59.65$ \\
        $k$-SVD, $k=50$ & $62.54$ & $55.01$ & $70.47$ & $70.70$ & $57.66$ & $25.84$ & $69.80$ & $58.09$ & $0.34$ & $0.17$ & $60.58$ \\
        $k$-SVD, $k=20$ & $62.41$ & $55.53$ & $70.60$ & $71.16$ & $58.76$ & $25.20$ & $70.40$ & $60.26$ & $0.33$ & $0.17$ & $59.94$ \\
        $k$-SVD, $k=15$ & $62.91$ & $55.48$ & $70.63$ & $71.25$ & $57.33$ & $23.89$ & $70.20$ & $59.71$ & $0.33$ & $0.16$ & $61.04$ \\
        \hline
        $k$-Sparse, $k=100$ & $65.22$ & $56.09$ & $\mathbf{76.47}$ & $72.04$ & $58.98$ & $27.69$ & $72.80$ & $70.33$ & $0.56$ & $0.37$ & $66.72$ \\
        $k$-Sparse, $k=50$ & $64.64$ & $57.13$ & $74.74$ & $71.51$ & $59.86$ & $27.42$ & $73.80$ & $71.36$ & $0.55$ & $0.32$ & $66.38$ \\
        $k$-Sparse, $k=20$ & $64.53$ & $55.98$ & $73.00$ & $71.65$ & $58.43$ & $26.24$ & $75.60$ & $68.20$ & $0.50$ & $0.25$ & $67.48$ \\
        $k$-Sparse, $k=15$ & $\mathbf{67.63}$ & $58.24$ & $75.52$ & $71.87$ & $\mathbf{62.16}$ & $\mathbf{30.14}$ & $76.80$ & $72.50$ & $0.55$ & $0.23$ & $65.45$ \\
        Sparsemax, $\tau=50$ & $64.58$ & $54.60$ & $66.33$ & $69.13$ & $55.46$ & $27.69$ & $65.80$ & $64.28$ & $0.53$ & $0.19$ & $\mathbf{67.88}$ \\
        Sparsemax, $\tau=20$ & $63.58$ & $54.58$ & $66.45$ & $70.21$ & $52.94$ & $26.92$ & $64.20$ & $63.95$ & $0.49$ & $0.19$ & $66.90$ \\
        Sparsemax, $\tau=10$ & $63.44$ & $54.60$ & $63.07$ & $69.29$ & $54.53$ & $26.92$ & $61.40$ & $63.02$ & $0.48$ & $0.17$ & $66.49$ \\
    \end{tabular}

    \end{subfigure}
    \caption{Evaluation of the embeddings from Movie-Dialogs models on various downstream tasks.
        The values measure classification accuracy or spearman correlation with human-labeled ground truth (see
        Section~\ref{sec:downstream}); larger values are better.}
    \label{tab:senteval}
\end{table*}

Additionally to the interpretability properties of sparse sentence embeddings, it is of interest whether
sparsity decreases their usefulness in downstream tasks. To evaluate this, we use the \emph{SentEval}
framework~\citep{senteval}, which learns downstream models on top of the provided sentence embeddings
to solve a variety of transfer tasks.

We report the accuracy on the standard classification problems the framework provides, namely
binary sentiment of movie reviews (MR), movie lines (SST2) and product reviews (CR), five-class sentiment of movie lines
(SST5), subjectivity/objectivity (SUBJ), binary opinion polarity (MPQA), and six-class question type (TREC) classification.
To look at semantic entailment/similarity of pairs of sentences, we report Spearman correlation with human-labelled
ground truth on the five-class semantic relatedness (STS14, SICK-R), and three-class semantic entailment (SICK-E) tasks,
and accuracy on the binary paraphrase detection (MRPC) task.

We configure the framework to use Logistic Regression for downstream models.
More details on the tasks and evaluation methods can be found in the SentEval paper.
We evaluate these tasks on the Movie-Dialogs models only, because COCO is unsuited for general-purpose sentence
embeddings.

We show the results of this evaluation in Table~\ref{tab:senteval}.
We again observe that sparse representations perform, in many cases, worse than their dense equivalent, therefore,
trading quality for interpretability\footnote{%
On a side note, we address the noticeable fact that the transfer tasks are solved with low accuracy in general.
For numbers comparable to the state of the art literature, more powerful sentence embedding models
\citecomment{such as self-attentive networks, InferSent, SkipThought etc., see}{infersent,skipthought}{}
with a higher latent dimensionality, more layers, and a larger and more diverse dataset are required.
Further, SentEval provides slower but more powerful MLP downstream models instead of Logistic Regression.}.
However, this does not occur on all tasks: for example, SST2 and SST5 clearly benefit from a sparse representation.

\subsection{Discussion}

The results of our quantitative evaluation method confirm the tendencies we observed in our empirical evaluation.
It appears that embedding dimensions generated by sparse models are coherent to a higher extent---in particular, the
lower the sparsity level, the more apparent topics can be found in the embedding dimensions.

The price of good interpretability is a higher reconstruction error.
As we impose more sparsity in the representation, the model is forced to ``cut corners'' and single slots in the embedding
are designated broader collections of traits in the data.
This results in more coherent topics, however, the narrow information bandwidth reduces the decoder net's
ability to reconstruct the exact sentence.
The fact that sparse representations carry less information may also explain the lower utility in some of the downstream tasks.
Other tasks (e.g.\ sentiment classification) can be solved with greater accuracy, which suggests that a sparse and
interpretable representation discovers more useful features for a simple downstream model like Logistic Regression.

As we force a model to deal with a sparse representation already during the training phase, finding sensible atoms
gets incorporated into the encoding and decoding mechanism.
We found that, in comparison to extracting this information from a dense model's intermediate
representation, this results in an observable and measurable boost in interpretability.
We can explain this by the fact that this architecture makes it \emph{part of the model's task} to find a sparse and
accurate representation of the data, whereas in a post-processing approach the model focuses on reconstruction only.

On the other side of the coin, this modification interferes with the training process.
To be more precise, the model follows a more complex objective, and the sparse layer is limited in the amount of
information that can be forward and backpropagated through it at a time---hence we observe convergence at a higher
loss value and bigger reconstruction errors.

We note that our Sparsemax-based approach does not perform particularly well in our evaluation, although in some
cases it outperforms other approaches when we consider samples beyond the 10 highest-ranked.
This can be explained by the fact that the sparsity level is not fixed, and that due to the Sparsemax layer,
the embeddings $\e_i$ are valid probability distributions.
A high value $e_{i,d}$ does, therefore, not necessarily indicate a strong presence of feature $d$ in sample $i$, but also
a lack of other features.
On the other hand, Sparsemax is better at determining feature presence\slash absence in general, due to not being
constrained to find an exact number of features.

\section{Related Work}
\label{sec:relatedwork}

As aforesaid, there has been work in the NLP literature on the interpretability of word embeddings.
\citet{nnse} suggest that sparse embeddings can be linked to a disentangled, and thus interpretable, representations.
This idea is also applied in~\citep{arora,spowv,spine}, commonly by solving a sparse dictionary learning problem on top
of dense embeddings.
In the papers \citep{sun2016sparse,chen2017learning,oiwe}, the authors learn sparse word embeddings during the training
phase.
\citet{thoughtvectors} applies above-named approaches to image embeddings, and the intermediate representation of an
image captioning model.

\citet{ksparseae} define the $k$-Sparse autoencoder. They use a $k$-Sparse layer in a shallow autoencoder
trained on the MNIST and NORB datasets, focusing on unsupervised feature learning, improvement in classification
accuracy, and a fast alternative to sparse coding.
\citet{sparsemax} develop Sparsemax as an alternative to the Softmax layer that is able to output exactly zero
probabilities, their work is focused on classification problems and attention mechanisms.

Interpretability metrics are usually of interest for word embeddings, where the predominant evaluation method is the
word intrusion test~\citep{wordintrusion}.
Our interpretability metric is based on topic coherence~\citep{topiccoherence}, a comparison of different variants of
this method can be found in~\citep{exploringtc}.

\section{Conclusion}
\label{sec:conclusion}

Being able to understand the intermediate representation of a neural net increases our model understanding.
In this paper we have taken a step towards this goal by introducing several sparse methods for a sentence autoencoder,
inspired by previous work on word embeddings.
The evaluation of our proposed models supports our hypothesis that sparse methods benefit the interpretability of the
embedding.
It is intuitive that a vector restricted to many zero values inevitably carries less information,
and indeed we have found that this increase in interpretability comes at a cost in reconstruction quality and, in some
cases, utility in downstream tasks.
It is, however, possible to strike a balance and achieve good interpretability without a large penalty.

We have devised a novel, automated method of quantifying said interpretability, based on topic coherence.
In our experiments, we observe that this evaluation corresponds to our manual assessment of interpretability.
It is fully automated, and therefore cheap and fast to run.
It can easily be extended by using different sentence similarity metrics or other topic coherence variants.

An interpretable sentence representation has further applications beyond model understanding: for example, it allows us
to develop a sentence similarity measure, that can justify \emph{why} two sentences are similar.
It can also help us understand downstream models on top of sentence embeddings.
For example, consider the case of a linear classification model:
we can inspect the largest positive and negative weights and understand which features in a source sentence influence
the model's decision.

For future work, it suggests itself to apply sparsity constraints to more sophisticated sentence embedding models
such as \emph{SkipThought} or \emph{InferSent}~\citecomment{respectively}{skipthought,infersent}{}.
Our methods can also be used to construct sparse encoder-decoder models for further tasks, such as image captioning,
machine translation, or recommender systems.

\bibliographystyle{acl_natbib_nourl}
\bibliography{refs}

\begin{thebibliography}{26}
\expandafter\ifx\csname natexlab\endcsname\relax\def\natexlab#1{#1}\fi

\bibitem[{Abadi et~al.(2015)Abadi, Agarwal, Barham, Brevdo, Chen, Citro,
  Corrado, Davis, Dean, Devin, Ghemawat, Goodfellow, Harp, Irving, Isard, Jia,
  Jozefowicz, Kaiser, Kudlur, Levenberg, Man\'{e}, Monga, Moore, Murray, Olah,
  Schuster, Shlens, Steiner, Sutskever, Talwar, Tucker, Vanhoucke, Vasudevan,
  Vi\'{e}gas, Vinyals, Warden, Wattenberg, Wicke, Yu, and Zheng}]{tensorflow}
Mart\'{\i}n Abadi, Ashish Agarwal, Paul Barham, Eugene Brevdo, Zhifeng Chen,
  Craig Citro, Greg~S. Corrado, Andy Davis, Jeffrey Dean, Matthieu Devin,
  Sanjay Ghemawat, Ian Goodfellow, Andrew Harp, Geoffrey Irving, Michael Isard,
  Yangqing Jia, Rafal Jozefowicz, Lukasz Kaiser, Manjunath Kudlur, Josh
  Levenberg, Dandelion Man\'{e}, Rajat Monga, Sherry Moore, Derek Murray, Chris
  Olah, Mike Schuster, Jonathon Shlens, Benoit Steiner, Ilya Sutskever, Kunal
  Talwar, Paul Tucker, Vincent Vanhoucke, Vijay Vasudevan, Fernanda Vi\'{e}gas,
  Oriol Vinyals, Pete Warden, Martin Wattenberg, Martin Wicke, Yuan Yu, and
  Xiaoqiang Zheng. 2015.
\newblock {TensorFlow}: Large-scale machine learning on heterogeneous systems.
\newblock Software available from tensorflow.org.

\bibitem[{Aharon et~al.(2006)Aharon, Elad, and Bruckstein}]{ksvd}
Michal Aharon, Michael Elad, and Alfred Bruckstein. 2006.
\newblock $ rm k $-svd: An algorithm for designing overcomplete dictionaries
  for sparse representation.
\newblock \emph{IEEE Transactions on signal processing}, 54(11):4311--4322.

\bibitem[{Arora et~al.(2016)Arora, Li, Liang, Ma, and Risteski}]{arora}
Sanjeev Arora, Yuanzhi Li, Yingyu Liang, Tengyu Ma, and Andrej Risteski. 2016.
\newblock Linear algebraic structure of word senses, with applications to
  polysemy.
\newblock \emph{arXiv preprint arXiv:1601.03764}.

\bibitem[{Chang et~al.(2009)Chang, Gerrish, Wang, Boyd-Graber, and
  Blei}]{wordintrusion}
Jonathan Chang, Sean Gerrish, Chong Wang, Jordan~L Boyd-Graber, and David~M
  Blei. 2009.
\newblock Reading tea leaves: How humans interpret topic models.
\newblock In \emph{Advances in neural information processing systems}, pages
  288--296.

\bibitem[{Chen et~al.(2017)Chen, Li, and Jin}]{chen2017learning}
Yunchuan Chen, Ge~Li, and Zhi Jin. 2017.
\newblock Learning sparse overcomplete word vectors without intermediate dense
  representations.
\newblock In \emph{International Conference on Knowledge Science, Engineering
  and Management}, pages 3--15. Springer.

\bibitem[{Cho et~al.(2014)Cho, Van~Merri{\"e}nboer, Gulcehre, Bahdanau,
  Bougares, Schwenk, and Bengio}]{gru}
Kyunghyun Cho, Bart Van~Merri{\"e}nboer, Caglar Gulcehre, Dzmitry Bahdanau,
  Fethi Bougares, Holger Schwenk, and Yoshua Bengio. 2014.
\newblock Learning phrase representations using rnn encoder-decoder for
  statistical machine translation.
\newblock \emph{arXiv preprint arXiv:1406.1078}.

\bibitem[{Conneau and Kiela(2018)}]{senteval}
Alexis Conneau and Douwe Kiela. 2018.
\newblock Senteval: An evaluation toolkit for universal sentence
  representations.
\newblock \emph{arXiv preprint arXiv:1803.05449}.

\bibitem[{Conneau et~al.(2017)Conneau, Kiela, Schwenk, Barrault, and
  Bordes}]{infersent}
Alexis Conneau, Douwe Kiela, Holger Schwenk, Loic Barrault, and Antoine Bordes.
  2017.
\newblock Supervised learning of universal sentence representations from
  natural language inference data.
\newblock \emph{arXiv preprint arXiv:1705.02364}.

\bibitem[{Danescu-Niculescu-Mizil and Lee(2011)}]{cornell}
Cristian Danescu-Niculescu-Mizil and Lillian Lee. 2011.
\newblock Chameleons in imagined conversations: A new approach to understanding
  coordination of linguistic style in dialogs.
\newblock In \emph{Proceedings of the Workshop on Cognitive Modeling and
  Computational Linguistics, ACL 2011}.

\bibitem[{Faruqui et~al.(2015)Faruqui, Tsvetkov, Yogatama, Dyer, and
  Smith}]{spowv}
Manaal Faruqui, Yulia Tsvetkov, Dani Yogatama, Chris Dyer, and Noah Smith.
  2015.
\newblock Sparse overcomplete word vector representations.
\newblock \emph{arXiv preprint arXiv:1506.02004}.

\bibitem[{Goh(2016)}]{thoughtvectors}
Gabriel Goh. 2016.
\newblock Decoding the thought vector.
\newblock http://gabgoh.github.io/ThoughtVectors/.

\bibitem[{Kingma and Welling(2013)}]{vae}
Diederik~P Kingma and Max Welling. 2013.
\newblock Auto-encoding variational bayes.
\newblock \emph{arXiv preprint arXiv:1312.6114}.

\bibitem[{Kiros et~al.(2015)Kiros, Zhu, Salakhutdinov, Zemel, Urtasun,
  Torralba, and Fidler}]{skipthought}
Ryan Kiros, Yukun Zhu, Ruslan~R Salakhutdinov, Richard Zemel, Raquel Urtasun,
  Antonio Torralba, and Sanja Fidler. 2015.
\newblock Skip-thought vectors.
\newblock In \emph{Advances in neural information processing systems}, pages
  3294--3302.

\bibitem[{Kusner et~al.(2015)Kusner, Sun, Kolkin, and Weinberger}]{wmd}
Matt Kusner, Yu~Sun, Nicholas Kolkin, and Kilian Weinberger. 2015.
\newblock From word embeddings to document distances.
\newblock In \emph{International Conference on Machine Learning}, pages
  957--966.

\bibitem[{Lin et~al.(2014)Lin, Maire, Belongie, Hays, Perona, Ramanan,
  Doll{\'a}r, and Zitnick}]{coco}
Tsung-Yi Lin, Michael Maire, Serge Belongie, James Hays, Pietro Perona, Deva
  Ramanan, Piotr Doll{\'a}r, and C~Lawrence Zitnick. 2014.
\newblock Microsoft coco: Common objects in context.
\newblock In \emph{European conference on computer vision}, pages 740--755.
  Springer.

\bibitem[{Luo et~al.(2015)Luo, Liu, Luan, and Sun}]{oiwe}
Hongyin Luo, Zhiyuan Liu, Huanbo Luan, and Maosong Sun. 2015.
\newblock Online learning of interpretable word embeddings.
\newblock In \emph{Proceedings of the 2015 Conference on Empirical Methods in
  Natural Language Processing}, pages 1687--1692.

\bibitem[{Makhzani and Frey(2013)}]{ksparseae}
Alireza Makhzani and Brendan Frey. 2013.
\newblock K-sparse autoencoders.
\newblock \emph{arXiv preprint arXiv:1312.5663}.

\bibitem[{Martins and Astudillo(2016)}]{sparsemax}
Andre Martins and Ramon Astudillo. 2016.
\newblock From softmax to sparsemax: A sparse model of attention and
  multi-label classification.
\newblock In \emph{International Conference on Machine Learning}, pages
  1614--1623.

\bibitem[{Mimno et~al.(2011)Mimno, Wallach, Talley, Leenders, and
  McCallum}]{mimno2011optimizing}
David Mimno, Hanna~M Wallach, Edmund Talley, Miriam Leenders, and Andrew
  McCallum. 2011.
\newblock Optimizing semantic coherence in topic models.
\newblock In \emph{Proceedings of the conference on empirical methods in
  natural language processing}, pages 262--272. Association for Computational
  Linguistics.

\bibitem[{Murphy et~al.(2012)Murphy, Talukdar, and Mitchell}]{nnse}
Brian Murphy, Partha Talukdar, and Tom Mitchell. 2012.
\newblock Learning effective and interpretable semantic models using
  non-negative sparse embedding.
\newblock \emph{Proceedings of COLING 2012}, pages 1933--1950.

\bibitem[{Newman et~al.(2010)Newman, Lau, Grieser, and
  Baldwin}]{topiccoherence}
David Newman, Jey~Han Lau, Karl Grieser, and Timothy Baldwin. 2010.
\newblock Automatic evaluation of topic coherence.
\newblock In \emph{Human Language Technologies: The 2010 Annual Conference of
  the North American Chapter of the Association for Computational Linguistics},
  pages 100--108. Association for Computational Linguistics.

\bibitem[{R{\"o}der et~al.(2015)R{\"o}der, Both, and Hinneburg}]{exploringtc}
Michael R{\"o}der, Andreas Both, and Alexander Hinneburg. 2015.
\newblock Exploring the space of topic coherence measures.
\newblock In \emph{Proceedings of the eighth ACM international conference on
  Web search and data mining}, pages 399--408. ACM.

\bibitem[{Subramanian et~al.(2017)Subramanian, Pruthi, Jhamtani,
  Berg-Kirkpatrick, and Hovy}]{spine}
Anant Subramanian, Danish Pruthi, Harsh Jhamtani, Taylor Berg-Kirkpatrick, and
  Eduard Hovy. 2017.
\newblock Spine: Sparse interpretable neural embeddings.
\newblock \emph{arXiv preprint arXiv:1711.08792}.

\bibitem[{Sun et~al.(2016)Sun, Guo, Lan, Xu, and Cheng}]{sun2016sparse}
Fei Sun, Jiafeng Guo, Yanyan Lan, Jun Xu, and Xueqi Cheng. 2016.
\newblock Sparse word embeddings using l1 regularized online learning.
\newblock In \emph{Proceedings of the Twenty-Fifth International Joint
  Conference on Artificial Intelligence}, pages 2915--2921. AAAI Press.

\bibitem[{Sutskever et~al.(2014)Sutskever, Vinyals, and Le}]{seq2seq}
Ilya Sutskever, Oriol Vinyals, and Quoc~V Le. 2014.
\newblock Sequence to sequence learning with neural networks.
\newblock In \emph{Advances in neural information processing systems}, pages
  3104--3112.

\bibitem[{Vinyals et~al.(2015)Vinyals, Toshev, Bengio, and Erhan}]{showandtell}
Oriol Vinyals, Alexander Toshev, Samy Bengio, and Dumitru Erhan. 2015.
\newblock Show and tell: A neural image caption generator.
\newblock In \emph{Computer Vision and Pattern Recognition (CVPR), 2015 IEEE
  Conference on}, pages 3156--3164. IEEE.

\end{thebibliography}

\appendix
\begin{table*}[t!]
    \small
    \centering
    \begin{subtable}{\textwidth}

        \begin{tabularx}{\textwidth}{c X}
            $e_{i,d_4}$ & $x_i$ \\
            \hline

            $0.98$ & \texttt{a black cat drinking water out of a water faucet .} \\
            $0.98$ & \texttt{the boats are outside on the water sailing .} \\
            $0.98$ & \texttt{several cows drinking water from a water receptacle .} \\
            $0.98$ & \texttt{a boat speeds down open water spraying water behind it .} \\
            $0.98$ & \texttt{two elephants drink water out of a body of water} \\
            $0.98$ & \texttt{a large body of water covered with boats .} \\
            $0.98$ & \texttt{a person stands on water skis in the water .} \\
            $0.98$ & \texttt{a woman is on the water on water skis .} \\
            $0.98$ & \texttt{small boats on water with setting sun behind distant hills .} \\
            $0.98$ & \texttt{a power boat on a body of water with a large water spray behind .} \\
        \end{tabularx}
        \caption{$coh_\text{WMD}(d_4) = -2.21$}
    \end{subtable}
    \medskip

    \begin{subtable}{\textwidth}
        \begin{tabularx}{\textwidth}{c X}
            $e_{i,d_5}$ & $x_i$ \\
            \hline

            $0.99$ & \texttt{white vase holding holding an assortment of flowers} \\
            $0.99$ & \texttt{a man holding holding a tennis racquet on a tennis court .} \\
            $0.99$ & \texttt{a man holding holding a giant remote control .} \\
            $0.99$ & \texttt{two bears holding each other outside the surroundings .} \\
            $0.99$ & \texttt{snowboarder holding a pink board being hugged by man in costume .} \\
            $0.99$ & \texttt{baby holding a teething toy in his hand} \\
            $0.99$ & \texttt{a countertop holding a <unk> bowl across from a shelf holding stemware .} \\
            $0.99$ & \texttt{a bird holding a fish in it's mouth .} \\
            $0.99$ & \texttt{a oven holding two trays of food baking .} \\
            $0.99$ & \texttt{two glasses holding red wine sit on a piece of paper on a wooden surface .} \\
        \end{tabularx}
        \caption{$coh_\text{WMD}(d_5) = -2.77$}
    \end{subtable}
    \medskip

    \begin{subtable}{\textwidth}
        \begin{tabularx}{\textwidth}{c X}
            $e_{i,d_6}$ & $x_i$ \\
            \hline
            $0.97$ & \texttt{person riding their skateboard on the street with the cars .} \\
            $0.94$ & \texttt{person riding a skateboard while pushing a stroller} \\
            $0.94$ & \texttt{person riding a horse while the sun sets} \\
            $0.93$ & \texttt{person riding a horse while another horse stands in a field .} \\
            $0.92$ & \texttt{person riding a bicycle while walking two dogs .} \\
            $0.92$ & \texttt{person riding a four wheeler on a beach towards a bridge .} \\
            $0.91$ & \texttt{person riding an elephant as it crosses through a river .} \\
            $0.91$ & \texttt{person riding a horse along shore of a body of water .} \\
            $0.91$ & \texttt{a person riding their bike down a path to a gate with a stop sign .} \\
            $0.91$ & \texttt{person riding down snowy hill on a pair of skis} \\
        \end{tabularx}
        \caption{$coh_\text{WMD}(d_6) = -2.21$}
    \end{subtable}
    \medskip

    \begin{subtable}{\textwidth}
        \begin{tabularx}{\textwidth}{c X}
            $e_{i,d_7}$ & $x_i$ \\
            \hline
            $0.99$ & \texttt{there is a brown box on the toilet} \\
            $0.99$ & \texttt{there is a blender with a green mixture in it} \\
            $0.99$ & \texttt{there is a brown bear walking through the woods alone} \\
            $0.99$ & \texttt{there is a clock that is above the building doors} \\
            $0.99$ & \texttt{there is a clock inside of a curvy blue sculpture .} \\
            $0.99$ & \texttt{there is a truck that has something mounted on the top} \\
            $0.99$ & \texttt{there is a boy playing with a tie} \\
            $0.99$ & \texttt{there is a person playing a nintendo wii} \\
            $0.99$ & \texttt{there is a boy playing baseball at the base ball field} \\
            $0.99$ & \texttt{there is a clock on the wall between the two arches .} \\
        \end{tabularx}
        \caption{$coh_\text{WMD}(d_7) = -3.47$.
            Note that this dimension has low coherence because the common feature it brings out (\texttt{there is a})
            consists of stop words, which are not considered in our metrics.
        }
    \end{subtable}
    \medskip

    \begin{subtable}{\textwidth}
        \begin{tabularx}{\textwidth}{c X}
            $e_{i,d_8}$ & $x_i$ \\
            \hline
            $1.00$ & \texttt{a person is surfing on a shallow wave .} \\
            $1.00$ & \texttt{a person is surfing on a medium sized wave .} \\
            $1.00$ & \texttt{a person is surfing in a on a wave} \\
            $1.00$ & \texttt{a person is surfing on a wave in the ocean .} \\
            $1.00$ & \texttt{a person is surfing a huge wave while staying upright .} \\
            $1.00$ & \texttt{a person is surfing on the waves of an empty ocean .} \\
            $1.00$ & \texttt{a person is surfing on a board at the beach} \\
            $1.00$ & \texttt{a person is surfing in a wave pool .} \\
            $1.00$ & \texttt{a person is surfing on a wave at the beach} \\
            $1.00$ & \texttt{a person is surfing a wave on a surfboard .} \\
        \end{tabularx}
        \caption{$coh_\text{WMD}(d_8) = -1.47$}
    \end{subtable}

    \caption{Highest-ranked samples in a selection of dimensions of our $k$-Sparse, $k=15$ model, trained on the COCO
        Captions data, along with the coherence of the dimension.
    }
    \label{tab:appendixtopsamples}
\end{table*}

\end{document}